\documentclass[letterpaper]{article} % DO NOT CHANGE THIS
\usepackage{aaai25}  % DO NOT CHANGE THIS
\usepackage{times}  % DO NOT CHANGE THIS
\usepackage{helvet}  % DO NOT CHANGE THIS
\usepackage{courier}  % DO NOT CHANGE THIS
\usepackage[hyphens]{url}  % DO NOT CHANGE THIS
\usepackage{graphicx} % DO NOT CHANGE THIS
\usepackage{natbib}  % DO NOT CHANGE THIS AND DO NOT ADD ANY OPTIONS TO IT
\usepackage{caption} % DO NOT CHANGE THIS AND DO NOT ADD ANY OPTIONS TO IT
\usepackage{multirow}
\usepackage{subcaption}
\usepackage{graphicx}
\usepackage{amssymb}
\usepackage{algorithm}
\usepackage{amsmath,amsfonts}
\usepackage{array}
\usepackage{subcaption}
\usepackage{textcomp}
\usepackage{url}
\usepackage{verbatim}
\usepackage{graphicx}
\usepackage{makecell}
\usepackage{booktabs}
\usepackage{algpseudocode}
\usepackage{notoccite}
\usepackage{multirow}

\usepackage{diagbox}
\usepackage{newfloat}
\usepackage{listings}
\DeclareCaptionStyle{ruled}{labelfont=normalfont,labelsep=colon,strut=off} % DO NOT CHANGE THIS
\floatstyle{ruled}
\newfloat{listing}{tb}{lst}{}
\floatname{listing}{Listing}
\title{UniPCGC: Towards Practical Point Cloud Geometry Compression via an Efficient Unified Approach}
\author{
    Kangli Wang\textsuperscript{\rm 1}, 
    Wei Gao\textsuperscript{\rm 1,2}\thanks{Corresponding Author: Wei Gao. }\\
}
\affiliations{
    \textsuperscript{\rm 1}Guangdong Provincial Key Laboratory of Ultra High Definition Immersive Media Technology, Shenzhen Graduate School, Peking University, Shenzhen 518055, China \\
    \textsuperscript{\rm 2}Peng Cheng Laboratory, Shenzhen, China\\
    kangliwang@stu.pku.edu.cn, 
    gaowei262@pku.edu.cn
}

\usepackage{bibentry}
\begin{document}

\maketitle

\begin{abstract}
Learning-based point cloud compression methods have made significant progress in terms of performance. However, these methods still encounter challenges including high complexity, limited compression modes, and a lack of support for variable rate, which restrict the practical application of these methods. In order to promote the development of practical point cloud compression, we propose an efficient unified point cloud geometry compression framework, dubbed as UniPCGC. It is a lightweight framework that supports lossy compression, lossless compression, variable rate and variable complexity. First, we introduce the Uneven 8-Stage Lossless Coder (UELC) in the lossless mode, which allocates more computational complexity to groups with higher coding difficulty, and merges groups with lower coding difficulty. Second, Variable Rate and Complexity Module (VRCM) is achieved in the lossy mode through joint adoption of a rate modulation module and dynamic sparse convolution. Finally, through the dynamic combination of UELC and VRCM, we achieve lossy compression, lossless compression, variable rate and complexity within a unified framework. Compared to the previous state-of-the-art method, our method achieves a compression ratio (CR) gain of 8.1\% on lossless compression, and a Bjontegaard Delta Rate (BD-Rate) gain of 14.02\% on lossy compression, while also supporting variable rate and variable complexity.
\end{abstract}

% Uncomment the following to link to your code, datasets, an extended version or similar.
%
\begin{links}
    \link{Project Page}{https://uni-pcgc.github.io/}
    % \link{Datasets}{https://aaai.org/example/datasets}
    % \link{Extended version}{https://aaai.org/example/extended-version}
\end{links}

% \maketitle
\begin{figure}
    \centering
    \includegraphics[width=1.0\linewidth]{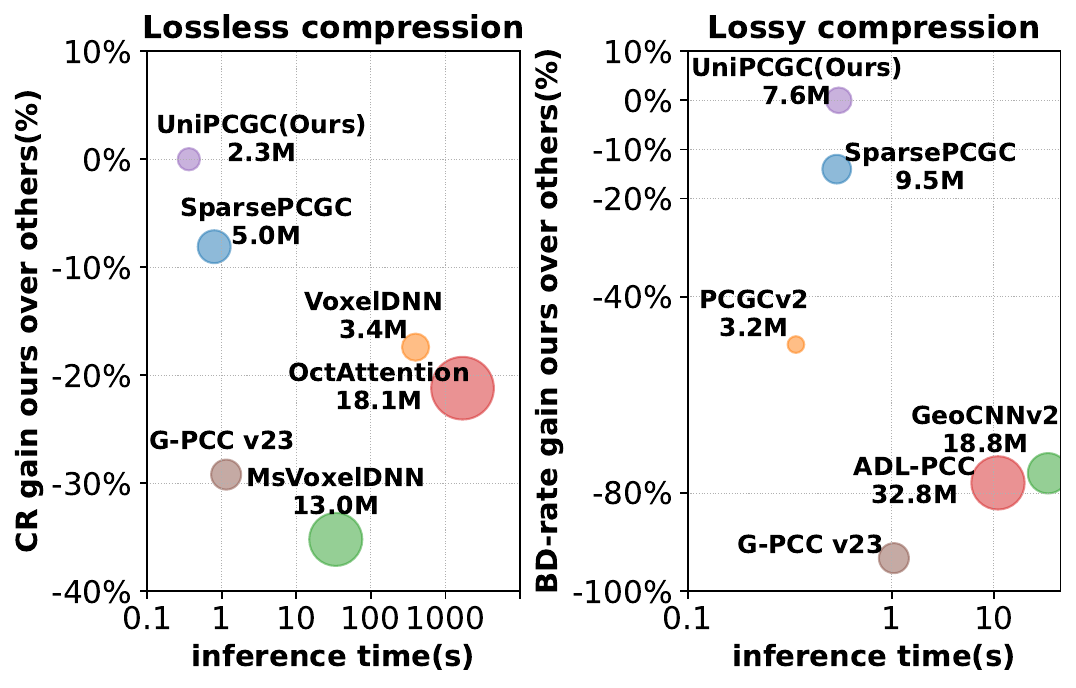}
    \caption{Performance comparisons for the proposed method and the other methods. The left figure shows the CR gain obtained by the proposed UniPCGC method in lossless compression. The right figure shows the BD-rate gain obtained by the proposed UniPCGC method in lossy compression.}
    \label{fig:xingneng}
\end{figure}
\section{Introduction}

3D point clouds have been widely used in applications such as immersive media communication and digital culture heritage, etc. There has been an increasing amount of point cloud data generated. To relieve the burden of massive data transmission and storage, there is an immediate requirement to develop point cloud compression algorithms with high efficiency. Although non-learning compression algorithm such as Geometry based Point Cloud Compression \cite{MPEG_PCC_PIEEE} (G-PCC) has been developed for point cloud compression, its performances is still not satisfactory. With the advancement of deep learning, numerous methods base on learning have been proposed. However, current learning-based methods still need to be improved in terms of performance and flexibility.

In point cloud compression, accurate probability estimation of point occupancy state can significantly reduce the bit rate. Some methods \cite{voxeldnn,msvoxel,fu2022octattention,cnet} achieve satisfactory performance by adopting autoregressive approaches for probability estimation. However, the decoding complexity of autoregressive-based methods is prohibitively high, making them impractical for real-world applications. Many efforts \cite{sparsepcgc,song2023efficient,ecm} focus on exploring more efficient grouping strategies to trade-off the performance and decoding complexity. However, the first group lacks any context information, which leads an unacceptably high bit rate. Reducing the bit rate of preceding groups becomes crucial for improving compression performance. 
To address this issue, we propose an uneven grouping scheme that reduces the size of the first group, allowing more points to benefit from context information. 

Flexibility is crucial for point cloud compression algorithms. In terms of flexibility, \cite{sparsepcgc} enhances flexibility by simultaneously supporting both lossy and lossless compression. In the field of image compression, certain lossy compression methods \cite{25,26} have achieved variable rate through modulation techniques. However, the paradigm may result in decreased compression performance under a wide range of variable rates. Overall, most existing works fall short in terms of supporting flexibility, highlighting the urgent need for a sufficiently flexible point cloud compression algorithm. To address the issue, we propose channel-level rate modulation and dynamic sparse convolution, which enable variable rate and complexity. 
The approach significantly enhances the flexibility and applicability of point cloud compression.

To address these limitations, we present an efficient unified framework named UniPCGC. In the lossless mode, we propose a method named Uneven 8-Stage Lossless Coder (UELC). In the lossy mode, Variable Rate and Complexity Module (VRCM) are achieved. By dynamically combining UELC and VRCM, we establish the UniPCGC framework. It is a lightweight framework that supports lossy compression, lossless compression, variable rate and variable complexity. As shown in Figure \ref{fig:xingneng}, we obtain state-of-the-art performance in both lossless and lossy mode. Moreover, our model exhibits competitive model size and inference speed. Our contributions are summarized as follows:

\begin{itemize}
\item We propose an efficient unified point cloud compression framework UniPCGC, which is a lightweight framework that supports lossy compression, lossless compression, variable rate and variable complexity. To our best knowledge, this is the first unified compression framework that simultaneously support these four modes.

\item To improve the efficiency of context grouping for better compression performances, we devise a method named Uneven 8-Stage Lossless Coder (UELC), which allocates more computational complexity to groups with higher coding difficulty, and merges groups with lower coding difficulty in the lossless mode.

\item To enhance the flexibility and applicability, we provide the Variable Rate and Complexity Module (VRCM), which are designed for lossy compression through the proposed joint adoption of a rate modulation scheme and dynamic sparse convolution.

% devise design develop provide propose present

\item Experimental results demonstrate that our method obtains state-of-the-art (SOTA) performance on both lossy and lossless compression, and supports variable rate and complexity, which facilitates practical applications.
\end{itemize}
\begin{figure*}
    \centering
    \includegraphics[width=0.84\linewidth]{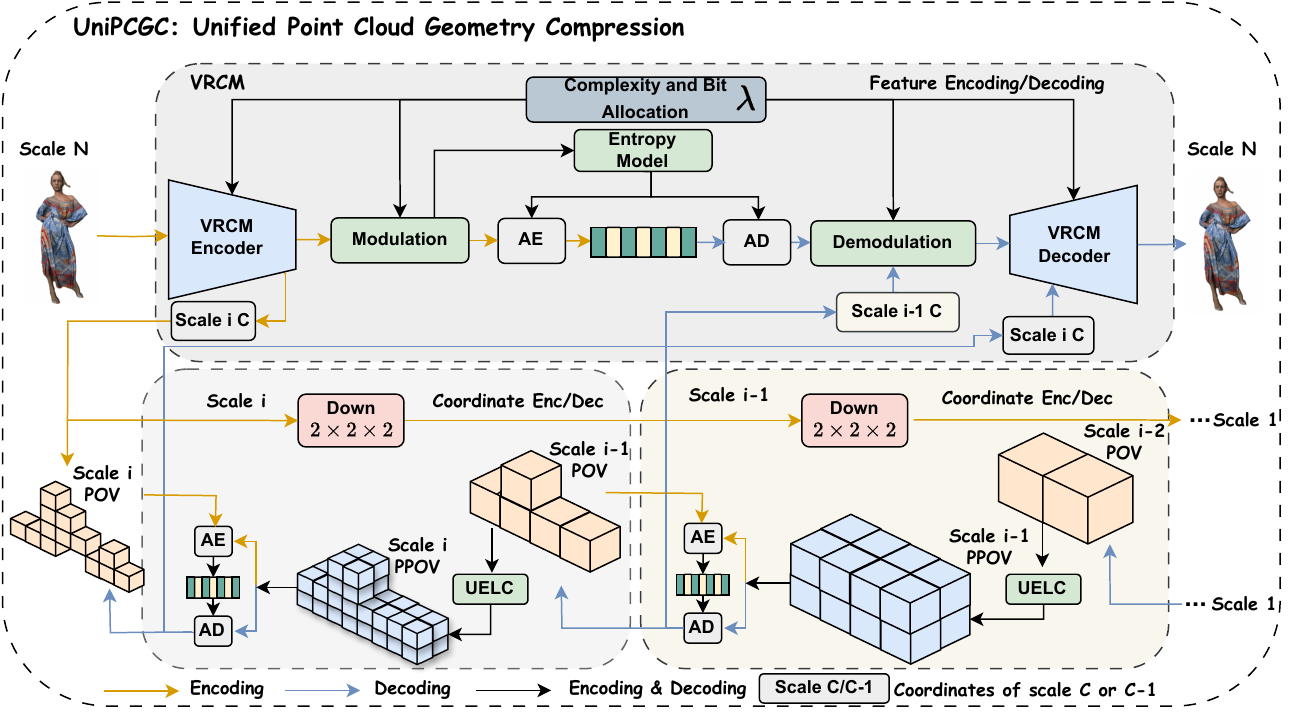}
    \caption{llustration of the proposed UniPCGC framework. It mainly consists of two parts: coordinate coding and feature coding. Coordinate coding is performed using Uneven 8-Stage Lossless Code (UELC) at each scale. Feature coding is performed using Variable Rate and Complexity Module (VRCM),  which mainly includes encoder, modulation, demodulation, decoder, complexity and bit allocation and Factorized Entropy Model. AE/AD stands for arithmetic encoding and arithmetic decoding.
    }
    \label{fig:zonghui}
\end{figure*}

\section{Related Works}

\subsection{Point Cloud Geometry Compression}
% \subsubsection{Voxel-Based Representation}

\textbf{Voxel-Based Representation.}
In the exploration of point cloud compression, various approaches have been proposed, including those based on 3D convolution neural networks. PCGCv1 \cite{pcgcv1} has shown promising rate-distortion efficiency. To achieve efficient and low-complexity compression, PCGCv2 \cite{pcgcv2} employs sparse convolution \cite{sparseconv} instead of traditional dense convolution. To utilize cross-scale information more efficiently, SparsePCGC has been proposed for both lossless and lossy compression, which estimates occupancy probabilities using cross-scale correlations in a single-stage or multi-stage manner \cite{sparsepcgc}. While these methods effectively maintain a relatively low level of complexity, their lack of support for variable rate and variable complexity limits their flexibility.
\\
% \subsubsection{Octree-Based Representation}
\textbf{Octree-Based Representation.}
Octsqueeze \cite{huang2020octsqueeze} and Muscle \cite{biswas2020muscle} adopt MLP for occupancy probability prediction. OctAttention \cite{fu2022octattention} uses an octree to represent point clouds to reduce spatial redundancy, and then designs a conditional entropy model to model ancestor and sibling contexts to exploit the strong dependencies between adjacent nodes. EHEM \cite{song2023efficient} resolves the issue of high decoding complexity by performing binary grouping of sibling nodes. Indeed, these methods encounter challenges in achieving satisfactory lossy compression performance on dense point clouds and fail to incorporate support for variable rate.
% \subsubsection{Hybrid Model}
\\
\textbf{Hybrid-Based Representation.} 
VoxelDNN \cite{voxeldnn} adaptively partitions point clouds into multi-resolution voxel blocks based on their structural characteristics and represents the partitioning using an octree. SparseVoxelDNN \cite{sparsevoxeldnn} replaces 3D convolution with 3D sparse convolution and achieves better performance based on VoxelDNN. Under the autoregressive framework, the decoding complexity will become very high. Such point cloud compression methods are difficult to apply in practical scenarios due to their high decoding complexity. GRASP-Net \cite{pang2022grasp} is another hybrid model framework. Although GRASP-Net maintains low complexity and achieves high performance, it only supports lossy compression and lacks support for variable rate, making it less practical in specific situations.
\subsection{Variable Rate Support}
$\lambda $ is also used to adjust the distortion weight and control the target rate, which means that different models need to be trained with different $\lambda$ values for different rate points, limiting the flexibility and generality of compression algorithms. To solve the problem, some methods use a single model to obtain variable rates. Conditional autoencoder schemes \cite{24,23,20,21,22} take prior information of $\lambda$ as input to obtain variable rates. \cite{27,26,25,28} define scaling factors or sub-networks to modify the magnitude of the latent representation, thereby obtaining variable rate through modulation. However, these works are primarily proposed in the field of image compression, and there is a scarcity of specifically tailored variable rate solutions available for point cloud compression.

\subsection{Dynamic Neural Networks} 
 % This development showcases the potential of dynamic networks in the compression domain. 
Dynamic networks utilizes a per-sample conditioned dynamic architecture for inference. It not only reduces redundant computations for simple samples but also maintains their representational capacity when dealing with challenging samples. 
% \subsubsection{Dynamic Depth}
Reducing redundant computations can be achieved through dynamic depth, which include early exiting \cite{35,37} and skip connections \cite{38,39,40}.
% \subsubsection{Dynamic Width} 
Dynamic width \cite{34,32,taoecp} refers to executing every layer and selectively skipping certain redundant channels to reduce computational complexity in a more fine-grained manner. 
Dynamic neural networks have been introduced in the field of image compression \cite{tao2023adanic}, enabling control the encoding complexity while achieving 40\% speedup. 
% This development showcases the potential of dynamic networks in the compression domain. 
However, there is a lack of available dynamic networks solutions specifically tailored for point cloud compression.

From the above literature review, we can see that although several methods have shown satisfactory performance, practical point cloud compression should ideally support lossless compression, lossy compression, variable rate and complexity simultaneously. It is noted that existing methods often consider only one or two of these aspects, which limits their applicability. To the best of our knowledge, the proposed UniPCGC is the first solution that achieves the unified support while obtaining superior compression performances.

% all four modes within a unified architecture and achieves state-of-the-art performance.
% in geometry compression of point clouds
\begin{figure*}
    \centering
    \includegraphics[width=0.95\linewidth]{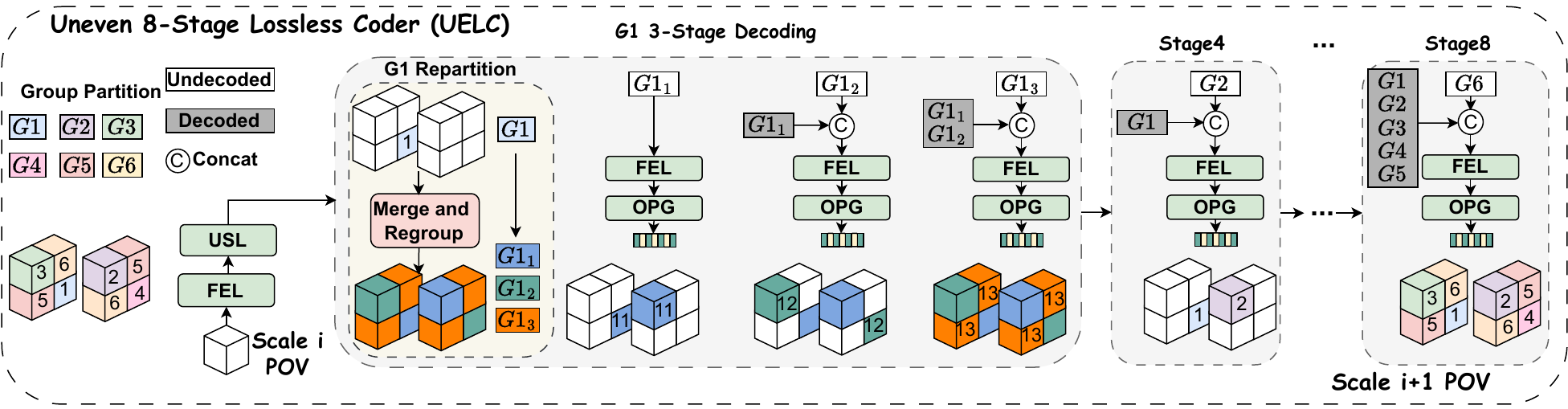}
    \caption{Illustration of the proposed Uneven 8-Stage Lossless Coder (UELC). In each stage, the previously coding groups and their features are regarded as prior information to better estimate the occupancy probability of groups in the current stage.}
    \label{fig:uneven_8stage_pcgc}
\end{figure*}
\section{Method}

\subsection{Overview}
The framework of the proposed UniPCGC is shown in Figure \ref{fig:zonghui}. UniPCGC mainly consists of two parts: coordinate coding and feature coding. In the lossless phase, the proposed Uneven 8-Stage Lossless Coder (UELC) is used. In the lossy phase, the proposed Variable Rate and Complexity Module (VRCM) is employed. UniPCGC enables lossless geometry compression by only using UELC for coordinate coding, while using UELC and setting dynamic downsampling times in VRCM allows lossy geometry compression at different rate ranges. 

\subsection{Notations and Blocks}
\textbf{Basic Notations.}
In our study, we represent point clouds using a voxelized approach. We denote a voxel as Positively or Non Occupied Voxel (POV or NOV) to indicate that the voxel is occupied or not occupied. 
PPOV represents Probable Positively Occupied Voxel. 
We utilize sparse convolution to extract features, where $k$ refers to the kernel size, $s$ refers to the stride, and $c$ refers to the number of channels. 
% Unless specified otherwise, the input and output channels remain consistent.
%This setting is similar to the one described in \cite{sparsepcgc}.
\\
\textbf{Feature Extraction Layer (FEL).}
We utilize the FEL for feature extraction. For dense object point clouds, sparse convolution with $k=3$ has proven effective in feature extraction. However, the distances within the groups increase with finer grouping. To address this, we expand the receptive field by setting $k=5$ to extract features. 
In FEL, we employ stacked Inception ResNet (IRN) for feature extraction.
% \subsubsection{USL: Upsampling Layer}
\\
\textbf{Upsampling Layer (USL).}
In the layer, we apply a transposed convolution with $k=2$ and $s=2$. 
% This operation allows one $POV$ to be refined into eight $PPOVs$.The proposed USL is illustrated in the Supplementary Material.
\\
\textbf{Occupancy Probability Generation (OPG).}
The layer is located after the FEL and utilized to calculate the occupancy probability for each voxel. The proposed FEL, IRN, USL and OPG are illustrated in the Supplementary Material.
% It typically comprises of three convolution layers followed by a sigmoid activation layer. The detailed architecture of OPG is illustrated in the Supplementary Material.
\\
\textbf{One-Stage Lossless Coder (OLC).}
OLC is commonly used for cross-scale probability estimation, utilizing the prior information of POV from $N-1$ scale to directly upsample to the $N$ scale in a single-stage. The detailed structure of OLC is illustrated in the Figure \ref{fig:vrc}.
\subsection{Uneven 8-Stage Lossless Coder (UELC)}

In SparsePCGC, a uniform eight-stage approach is proposed, which achieves SOTA performance in lossless geometry compression. Upon re-evaluating this grouping method, we observe that the first group lacks any reference information, leading to an excessive number of bits required for encoding. However, in the later groups, more refined grouping brings little benefit. Therefore, we believe that more coding stages should be allocated in the first group, and subsequent groups should be combined. Based on this motivation, we propose the Uneven 8-stage Lossless Coder (UELC). 
In lossless compression, UELC mainly performs the probability prediction of PPOV. As shown in Figure \ref{fig:zonghui}, the more accurate the probability of PPOV estimation after UELC, the smaller bits required for arithmetic encoding. At the decoding end, arithmetic decoding can be used to restore the current scale POV.
Figure \ref{fig:uneven_8stage_pcgc} details the UELC step by step:

\textbf{(1)} First, the POV of Scale $i$ is upsampled through USL and FEL to obtain the PPOV, which is divided into 6 groups as shown in the Figure \ref{fig:uneven_8stage_pcgc}. The processing within the group is parallel, and the processing between the groups is serial. After all processing is completed, we obtain the prediction of the occupancy probability of scale i+1.

\textbf{(2)} When encoding or decoding $G1$, we group $G1$ more precisely and divide $G1$ into $G1_1$, $G1_2$, and $G1_3$, marked as 11, 12, and 13 in Figure \ref{fig:uneven_8stage_pcgc}. Three groups are processed sequentially and complete in three stages. For each stage, we process the PPOV using stacked FEL and OPG blocks to determine its $p_\mathrm{PPOV}$ and compress the occupancy probability into the bitstream in the encoder or parse the bitstream in the decoder to identify the POV and NOV.

\textbf{(3)} For the current encoding or decoding stage, we prune the PPOV of the previous stage according to the POV, and use the pruned PPOV as prior information to assist in predicting the occupancy probability of the current stage.

\textbf{(4) }The subsequent stages are similar to the previous stage. We gradually complete the probability estimation for the entire scale using the proposed uneven 8-stage approach. The entire coordinate coding process can be expressed by the following formula:
\begin{equation}
\begin{aligned}
& \operatorname{bin}_i=f_E\left(C_{i-1}, C_i\right)=\left(\operatorname{AE}\left(f_{ {U }}\left(C_{i-1}\right), C_i\right)\right),  \\
& C_i=f_D\left(C_{i-1}, \operatorname{bin}_i\right)=\left(\operatorname{AD}\left(f_{ {U }}\left(C_{i-1}\right),  { bin }_i\right)\right), 
\end{aligned}
\end{equation}
where $f_U(\cdot)$ is the UELC, $C_i$ represents the coordinates of scale i, and $bin_i$ is the compressed file of scale i.
In addition, the proposed UELC employs a non-sequential grouping scheme. Since UELC requires merging certain groups for joint encoding, it is essential to ensure that these merged groups can reference information from closer points as much as possible. Inspired by checkerboard context \cite{checkerboard}, we propose a non-sequential grouping scheme, as shown in Figure \ref{fig:uneven_8stage_pcgc}.
At the m-th scale, we can estimate the bit rate of PPOV occupancy status:
%using the equation \eqref{eq:uelc}:
\begin{equation}
R_{\mathrm{S}}(m)=\sum_k-\log _2\left(p_{\mathrm{P}\mathrm{POV}}^m(k)\right), \label{eq:uelc}
\end{equation}
where $p_{\mathrm{P}\mathrm{POV}}^m(k)$ is the probability estimated by UELC. The aforementioned process involves occupancy probability estimation between two scales. We apply the proposed UELC from the lowest scale to the highest scale, ensuring that each scale shares the same weights, which enhances the flexibility of the model.

\subsection{Variable Rate and Complexity Support}
\begin{figure*}
    \centering
\includegraphics[width=0.95\linewidth]{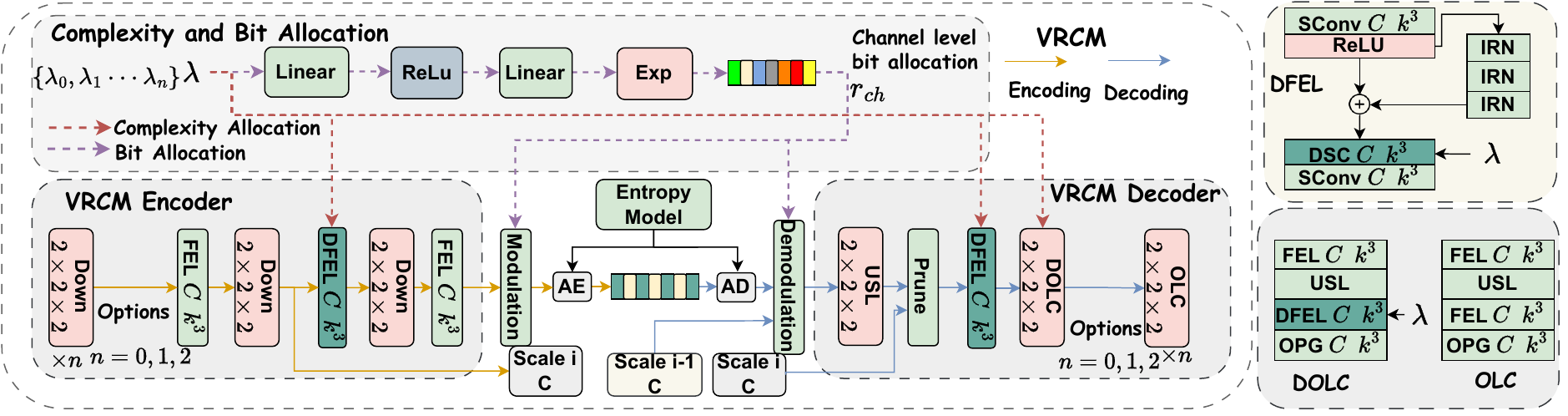}
    \caption{Detailed architecture of proposed VRCM. It also shows the architecture of channel level bit allocation module, One-Stage Lossless Coder (OLC / DOLC) and Dynamic Feature Extraction Layer (DFEL).}
    \label{fig:vrc}
\end{figure*}
In this section, we propose the Variable Rate and Complexity Module (VRCM) for lossy geometry compression to support variable rate and variable complexity. The Figure \ref{fig:vrc} illustrates the main framework of VRCM.
VRCM completes the feature encoding part, and the coordinate $C_i$ encoding is completed by UELC. Therefore, at the decoding end, $C_i$ and $C_{i-1}$ are prior information.

At the encoder, different downsampling times\textbf{ (0, 1, 2) }are performed according to the bit rate range (high, medium, low). It can be represented by $f_{down}^n(\cdot)$, where n is the number of downsampling times. Subsequently, the latent feature $y$ is obtained by passing through the FEL, Down, DFEL, Down, FEL layers respectively. After passing through the first Down layer, the coordinates of this scale $C_i$ are sent to the coordinate encoding part for lossless encoding. The above process can be represented by $f_{enc}(x,\lambda)$. Among them, $\lambda$ is the complexity and rate modulation factor, which modulates the computational complexity of the module at the encoder and decoder. Moreover, $\lambda$ generates the channel-level rate modulation coefficient through channel level bit allocation module, which is modulated with $y$ at the encoder and demodulated by $\hat{y}_m$ at the decoder. At the decoder, the demodulated features are combined with $C_{i-1}$ to form a sparse tensor. After USL, the correct coordinates $C_i$ will be used to crop the wrong voxels to reduce distortion. Finally, it passes through DFEL, DOLC and $f_{up}^n(\cdot)$. The above process can be expressed by the following formula:
\begin{equation}
\begin{gathered}
y=f_{ {enc }}\left(f_{ {down }}^n\left(p_{ {ori }}\right), \lambda\right), \\
r_{ {ch }}=f_{ {Exp }}\left(f_L\left(f_R\left(f_L(\lambda)\right)\right)\right), \\
y_m=y * r_{ {ch }}, \hat{y}{_m}=Q\left(y_m\right), y_{d m}=\hat{y}{_m} / r_{c h}, \\
\hat{y}={Cat}\left(y_{ {dm }}, C_{i-1}\right), \\
p_{ {res }}=f_{ {up }}^n\left(f_{ {dec }}\left(\hat{y}, \lambda, C_i\right)\right),
\end{gathered}
\end{equation}
where ${p_{ori}}$ is the original input point cloud, $f_L(\cdot)$ is the linear layer, $f_R(\cdot)$ is the Relu activation function, $f_{Exp}(\cdot)$ is the exponential function, $r_{ch}$ is the channel-level rate modulation coefficient, $C_i$ represents the coordinate of scale i, and $Cat(\cdot)$ means merging the input to generate a sparse tensor. The value of $R_{y_m}(i)$ can be calculated as follows:
\begin{equation}
R_{y_m}(i)=\sum_j-\log _2\left(p_{R_{y_m}^{i}}(j)\right) \label{eq:f},
\end{equation}
where $p_{R_{y_m}^{i}}(j)$ follows the factorized entropy model. 

Moreover, in the works of PCGCv2 and SparsePCGC, the factorized entropy model \cite{balle2018variational} is utilized. It assumes that there is no statistical dependency within the probability distribution. Hence, the essence of compression lies in removing statistical dependencies as much as possible during the encoding process \cite{ali2024towards}. Inspired by this, we propose dynamic sparse convolution (DSC), where points with high correlation are allocated more computational resources to explicitly remove statistical dependencies, thereby maintaining compression performance while reducing encoding complexity. 
We can define the correlation between the current point and surrounding points as:
\begin{equation}
{Corr}={E}_{x \sim p(x)}[(\frac{x_i-\mu_i}{\sigma_i})(\frac{x_c-\mu_c}{\sigma_c})],
\end{equation}
where $0 \leq i<k^2$, $c$ represents the current point, and $\mu$ and $\sigma$ represent the mean and variance of the feature vector of the points. Next, we divide the points in the space into two groups based on the correlation.
% By introducing the complexity control factor $f$, we can control the number of points involved in the computations, allowing for a fine-grained and adjustable computational complexity. 
The specific calculation definition is as follows:

\begin{equation}
\begin{gathered}
x_1, x_2=f_{ {spilt }}\left(f_{ {conv }}(x), \lambda \mid  { Corr }\right), \\
 { out }=f_{ {conv }}\left(x_1\right) \cup x_2,
\end{gathered}
\end{equation}
where $f_{spilt}(\cdot)$ is the split operation, $f_{conv}(\cdot)$ is the sparse convolution layer, and $\cup$ is the concatenation of two sparse tensors. Next, we replace the convolutions in OLC and FEL with DSC, resulting in Dynamic One-Stage Lossless Coder (DOLC) and Dynamic Feature Extraction Layer (DFEL), as shown in the Figure \ref{fig:vrc}.
% As shown in the Figure \ref{fig:vrc}, the features have been downsampled $n$ times are compressed into the bit stream. According to the equation \ref{eq:f}, we can calculate the value of $R_F(i)$. Subsequently, after $n-1$ upsampling operation, they are restored to their original scale using DOLC. It is worth noting that since the true coordinates after upsampling are provided by lossless compression, we can prune the point cloud to reduce distortion. Finally, the point cloud reconstructed through DOLC is adaptively pruned to the points number of the original scale, similar to the approach described in \cite{pcgcv2}. 
The above is the main introduction of VRCM. Next, we discuss the training methodology for VRCM. We propose a multi-stage training strategy for VRCM. \textbf{Stage1:} We set the $\lambda$ value of the loss function to $0.1$ to achieve fast optimized. \textbf{Stage2:} We set the $\lambda$ value to $1.0$ for stable optimized. \textbf{Stage3:} We train multiple loss functions with rate control factors $\lambda$ to achieve variable rate and variable complexity, as shown in equation \eqref{eq:loss_all_stage3}.
% \begin{algorithm}
% \caption{Multi-stage Training Strategy}\label{algorithm}
% % \KwData{current period $t$, initial inventory $I_{t-1}$, initial capital $B_{t-1}$, demand samples}
% % \KwResult{Optimal order quantity $Q^{\ast}_{t}$}
% Stage1:Quick optimization of single bit rate by set $\lambda=0.1$\;
% % \For{$epoch\leftarrow 0$ \KwTo $1$}{
% % % $\lambda \leftarrow 0.1,f \leftarrow 1.0$\;
% % $loss \leftarrow R(\hat{y} ; \theta, f)+0.1 × D(x, \hat{x} ; \theta, f) $\;
% % % $loss.backward()$\;
% % }
% Stage2:Stable optimization of single bit rate by set $\lambda=1.0$\;
% % \For{$epoch\leftarrow 0$ \KwTo $1$}{
% % $\lambda \leftarrow 1.0,f \leftarrow 1.0$\;
% % $loss \leftarrow R(\hat{y} ; \theta, f)+1.0 × D(x, \hat{x} ; \theta, f) $\;
% % % $loss.backward()$\;
% % }
% Stage3:Multi-rate training\;
% \For{$epoch\leftarrow 0$ \KwTo $20$}{
% $T \leftarrow \left\{f_0, f_1 \cdots f_n\right\}$\;
% $loss \leftarrow \sum_{f \in T}[R(\hat{y} ; \theta, f)+f D(x, \hat{x} ; \theta, f)]$\;
% % $loss.backward()$\;
% }
% \end{algorithm}
% \end{document}

\subsection{Loss Functions}
\begin{table*}[]
% \caption{Performance of lossless methods on the 8iVFB testset under the same training conditions. UniPCGC, GPCC v23 and SparsePCGC are tested using RTX 4080 GPU and intel i5-13600KF CPU for a fair runtime comparison (Mark with *).}
% \label{table:lossless_8i}
\centering
\begin{tabular}{|c|c|c|c|c|c|c|c|}
\hline
\textbf{Point clouds} & \textbf{Ours}  & \textbf{GPCC} & \textbf{SparsePCGC} & \textbf{VoxelDNN} & \textbf{MsVoxelDNN} & \textbf{OctAttention} & \textbf{ECM-OPCC} \\ \hline
Red\&black × 300            & 0.59           & 0.82         & 0.64                & 0.67         & 0.87            & 0.73                  & 0.66              \\
Loot × 300                 & 0.49           & 0.69         & 0.53                & 0.58         & 0.73            & 0.62                  & 0.55              \\
Thaidancer × 1            & 0.51           & 0.70         & 0.56                & 0.68         & 0.85            & 0.65                  & 0.58              \\
Boxer × 1                & 0.45           & 0.65         & 0.49                & 0.55          & 0.70             & 0.59                  & 0.51              \\ \hline
\textbf{Average Bpp } $\downarrow$ & \textbf{0.51}  & 0.72        & 0.56               & 0.62       & 0.79          & 0.65                & 0.58             \\ \hline
\textbf{Ours CR Gain} $\uparrow$ & \textbf{0.0\%} & -29.2\%       & -8.1\%              & -17.4\%       & -35.2\%         & -21.2\%               & -11.3\%           \\ \hline
\textbf{Enc time(s)} $\downarrow$  & *0.56   & *1.99          & *0.71                 & 885           & 54              & 1.06                  & 1.92              \\ \hline
\textbf{Dec time(s)} $\downarrow$  & *0.57  & *1.49          & *0.88                 & 640           & 58              & 1229                  & 19.5              \\ \hline
\end{tabular}
\caption{Performance of lossless methods on the 8iVFB testset under the same training conditions. UniPCGC, GPCC v23 and SparsePCGC are tested using RTX 4080 GPU and intel i5-13600KF CPU for a fair runtime comparison (Mark with *).}
\label{table:lossless_8i}
% \captionsetup{justification=raggedright,singlelinecheck=false,format=hang}
% \caption*{*Ours and SparsePCGC runtime test on RTX4080, while SparsePCGC and other mothods test on RTX2080}
% \caption*{*UniPCGC and SparsePCGC are tested on RTX4080}
%We directly cite the testing time from [20]. Due to the lack of the same GPU device (RTX 2080) as [20] for testing, we conduct the tests for UniPCGC and SparsePCGC using RTX 4080 to achieve a fair comparison.
\end{table*}
% \begin{figure*}[t]
% 	\centering
% 	% \subfloat{
% 	% \includegraphics[width=1.72in]{img/sub_rd_cur/loot_vox10_1200_d1.pdf}
% 	% \includegraphics[width=1.72in]{img/sub_rd_cur/loot_vox10_1200_d2.pdf}
% 	% \includegraphics[width=1.72in]{img/sub_rd_cur/queen_0200_d1.pdf}
% 	% \includegraphics[width=1.72in]{img/sub_rd_cur/queen_0200_d2.pdf}
% 	% }
%  \subfloat{
%     \includegraphics[scale=0.118]{img/sub_rd_cur/loot_vox10_1200_d1.pdf}
%     \includegraphics[scale=0.118]{img/sub_rd_cur/loot_vox10_1200_d2.pdf}
%     \includegraphics[scale=0.118]{img/sub_rd_cur/queen_0200_d1.pdf}
%     \includegraphics[scale=0.118]{img/sub_rd_cur/queen_0200_d2.pdf}
%     }
% % 	\\
% % 	\subfloat{
% % \includegraphics[width=1.62in]{img/exp/longdress_vox10_1300_d2.pdf}
% % 	\includegraphics[width=1.62in]{img/exp/loot_vox10_1200_d2.pdf}
% % 	\includegraphics[width=1.62in]{img/exp/queen_0200_d2.pdf}
% % 	\includegraphics[width=1.62in]{img/exp/basketball_player_vox11_d2.pdf}}
% 	\caption{{Performance comparison using rate-distortion curves. }}
% 	\label{fig:rdcurves}
% \end{figure*}
\begin{figure*}[t]
    \centering
    \begin{subfigure}{\textwidth}
        \includegraphics[scale=0.118]{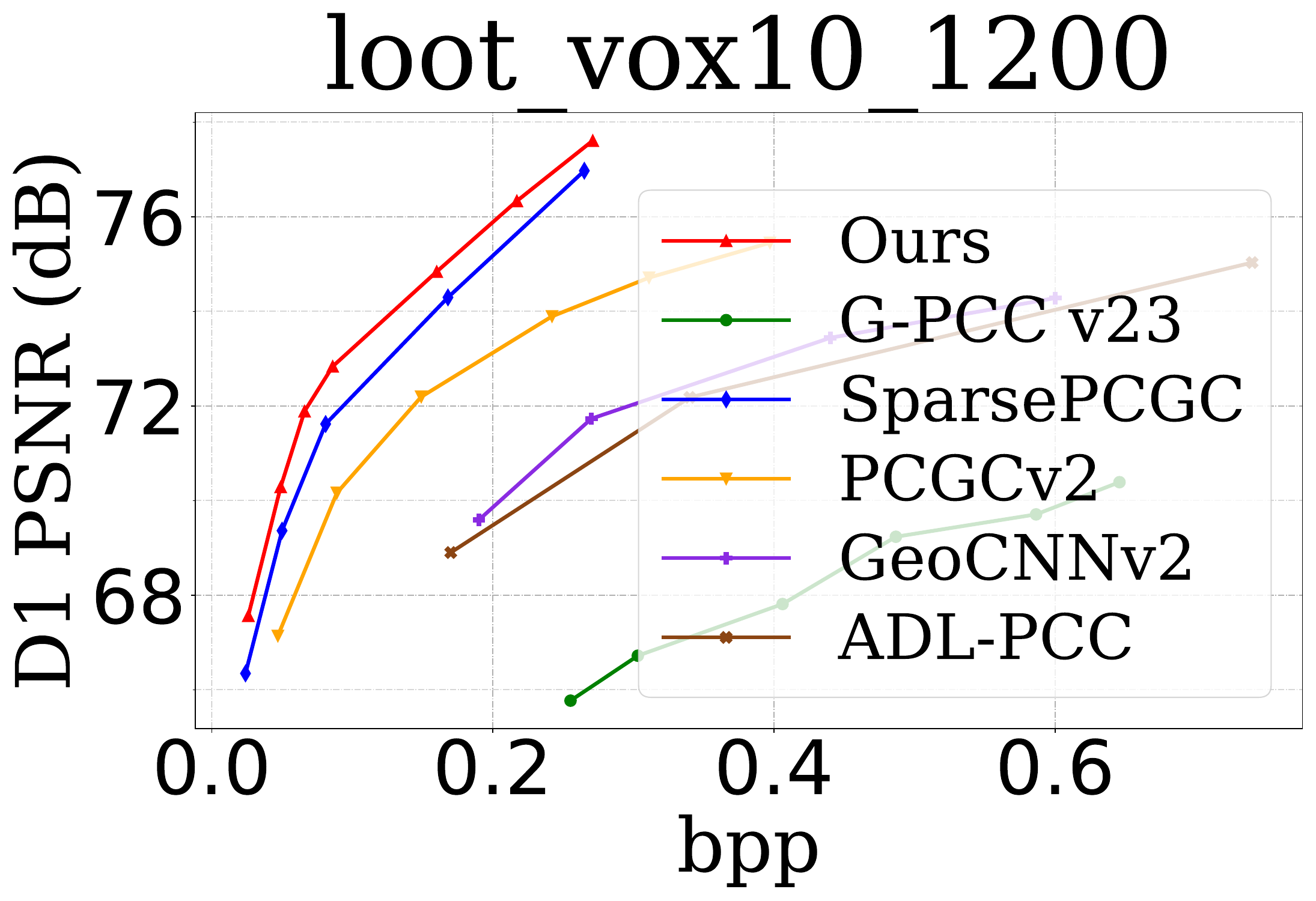}
        \includegraphics[scale=0.118]{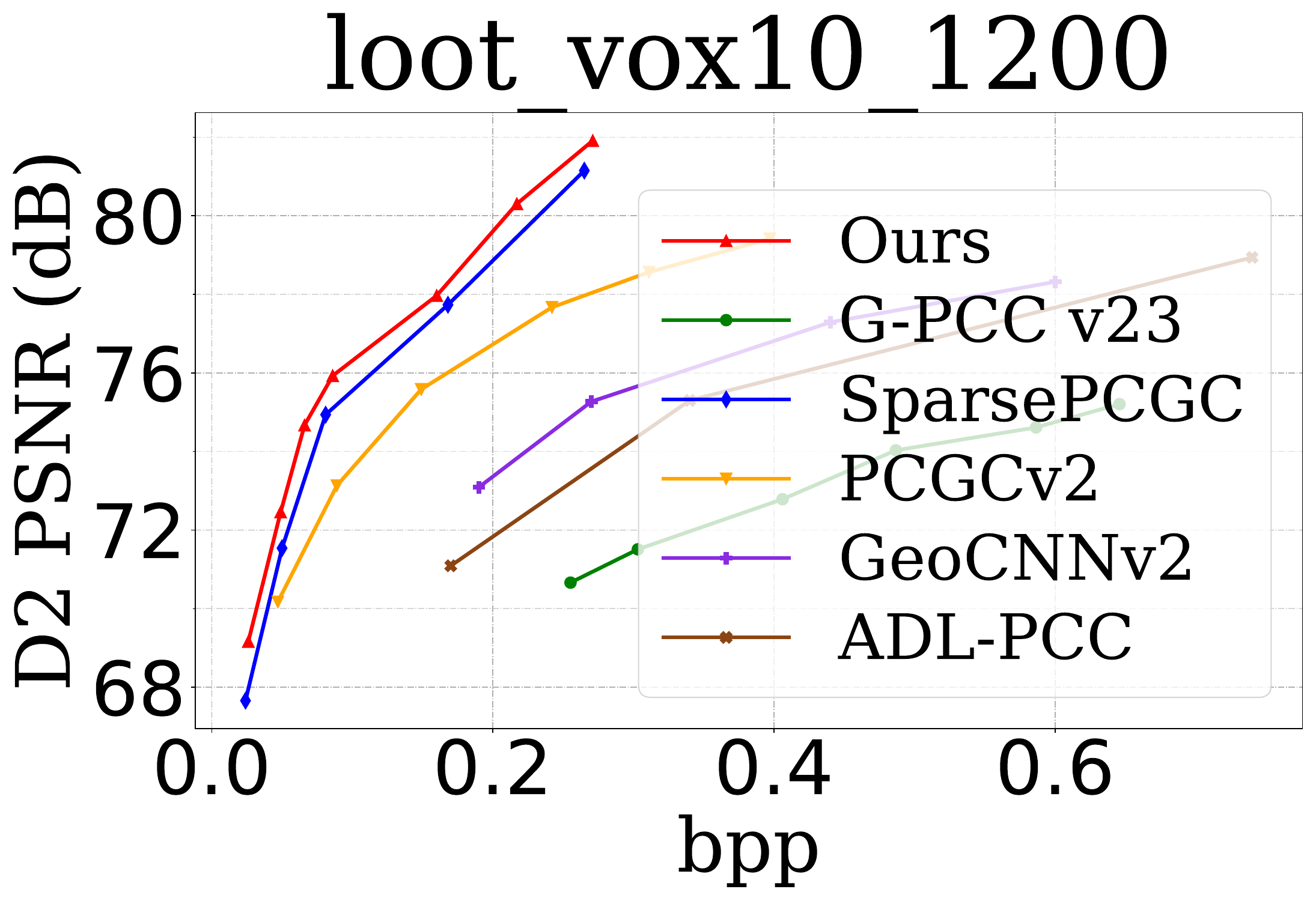}
        \includegraphics[scale=0.118]{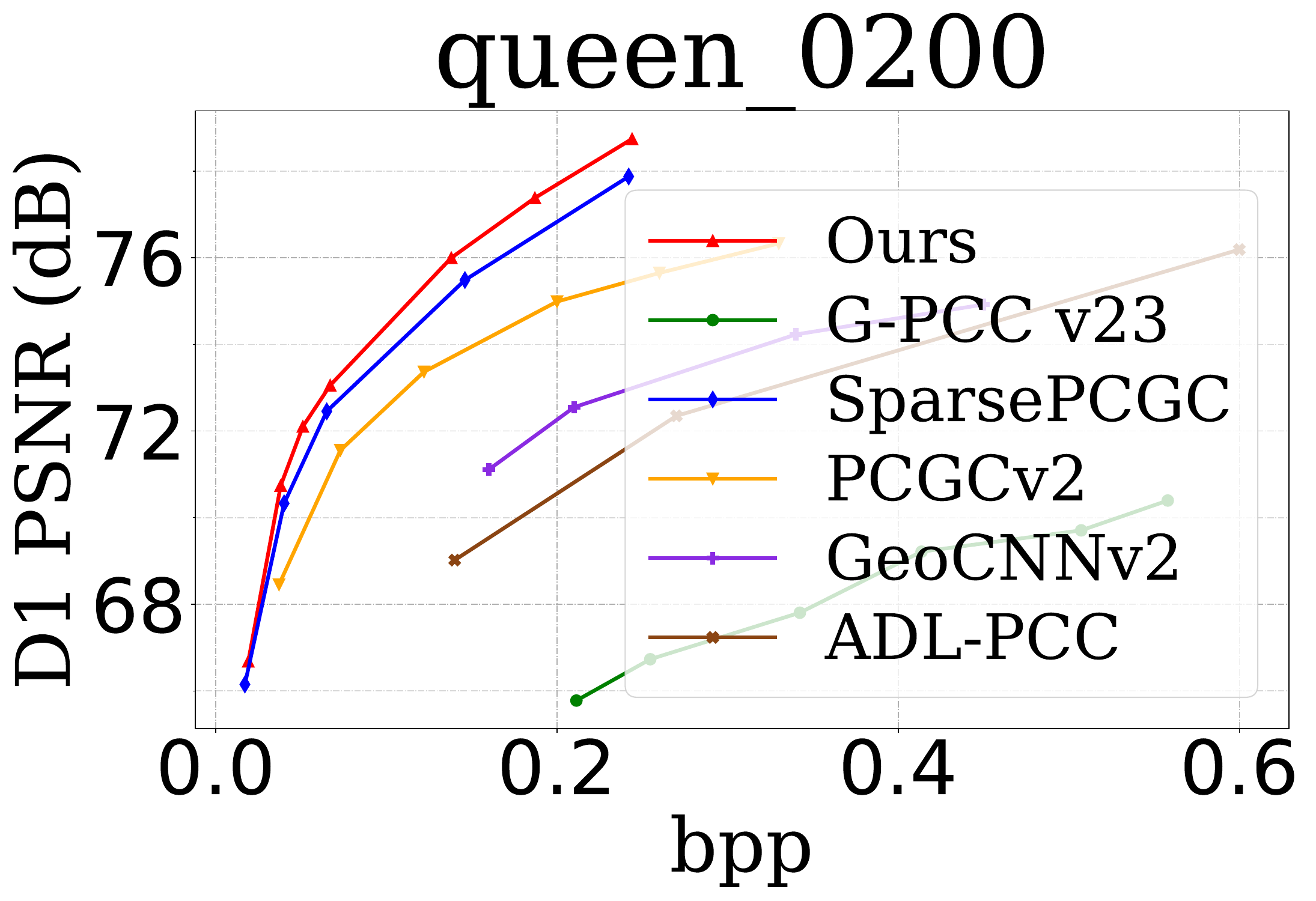}
        \includegraphics[scale=0.118]{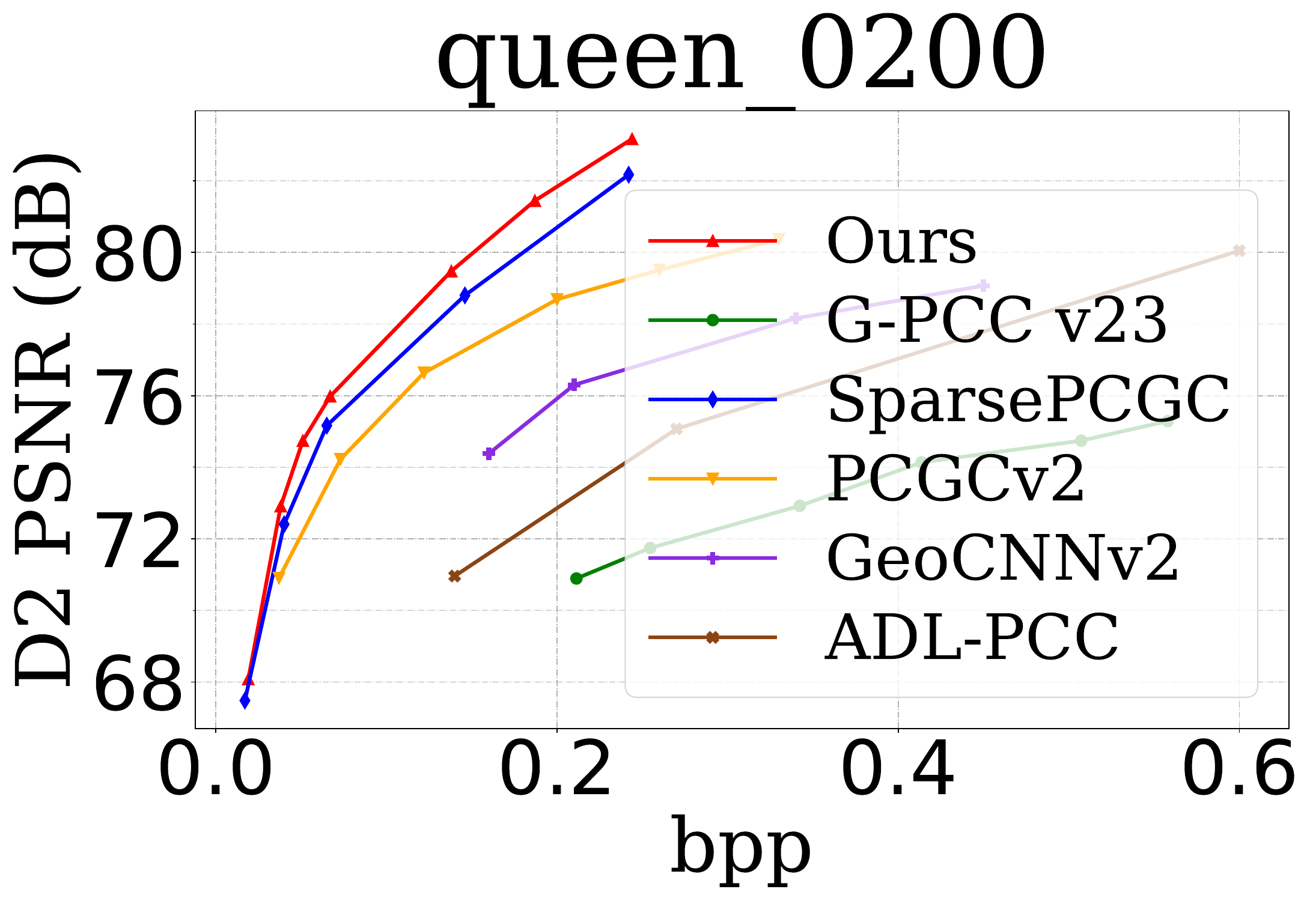}
    \end{subfigure}
    \caption{Performance comparison using rate-distortion curves.}
    \label{fig:rdcurves}
\end{figure*}
We use Binary Cross Entropy (BCE) in training to estimate the difference between predicted occupancy probabilities and actual occupancy labels:
\begin{equation}
L_{BCE}=\sum_k-(o_1(k) \log _2 p_1(k)+ o_2(k) \log _2 p_2(k)), \label{eq:loss}
\end{equation}
% \begin{equation}
% L_{B C E}=\sum_k-\left(o(k) \log _2(p(k))+(1-o(k)) \log _2(1-p(k))\right), \label{eq:loss}
% \end{equation}
where $o_1(k)$ and $o_2(k)$ represent the actual occupancy and non-occupancy symbol respectively, $p_1$ and $p_2$ are the probability that the k-th PPOV is POV or NOV. In lossless compression, we train UELC on the highest four scales. In lossy compression, we train VRCM with n set to 0, 1, and 2. Additionally, the loss function must incorporate the rate consumption of features, as shown in the following equation:
\begin{equation}
L_{{all}}=\lambda L_{{BCE}}+ R_{y_m},\label{eq:loss_all}
\end{equation}
where $R_{y_m}$ is calculated by equation \eqref{eq:f} and 
the above equation is used in the first and second stages of training in lossy compression. In the third stage, we train multiple loss functions with rate control factors $\lambda$ to achieve variable rate and variable complexity as follows:
\begin{equation}
L_{{all}}= \sum_{\lambda \in T}[R(\hat{y} ; \theta, \lambda)+\lambda L_{BCE}(x, \hat{x} ; \theta, \lambda)], 
 \label{eq:loss_all_stage3}
\end{equation}
where $T \in \left\{\lambda_0, \lambda_1 \cdots \lambda_n\right\}$, $T$ is the preset modulation factor set and we calculate the total loss to train models of different complexity and rate.

\section{Experimental Results}
% This section primarily describes the experimental setup and results of UniPCGC.
\subsection{Experiment Setup}

\subsubsection{Dataset.}
In our experiment, we use the ShapeNet dataset and the selected dense point clouds. ShapeNet \cite{chang2015shapenet} is a 3D object CAD model dataset, encompassing a subset known as ShapeNet-Core. The subset comprises 55 common object categories and approximately 51,300 unique 3D mesh models. We densely sample points on the original meshes to generate point clouds. Subsequently, we randomly rotate and quantize them with 8-bit geometry precision in each dimension. We use the processed shapenet dataset for training of the proposed UniPCGC. For dense point clouds, we choose from MPEG PCC dataset, including longdress 1300, redandblack 1550, soldier 0690, loot 1200, queen 0200, basketball player 0200, and dancer 0001. We select these point cloud as testset for lossy and lossless compression. We also employ the 8iVFB and MVUB sequences for training and 8iVFB sequences for testing to align with popular lossless compression methods.
% \subsubsection{ShapeNet}
% \subsubsection{Dense Point Clouds}
\subsubsection{Training Strategies.}
We use MinkowskiEngine \cite{sparseconv} and Pytorch to build our model, and perform UELC and VRCM training on RTX4080 GPU and intel i5-13600KF CPU. For UELC, it is trained for lossless compression using the loss function equation \eqref{eq:loss}. We initialize the learning rate to $8 \times 10^{-4}$ and gradually decrease it to $5 \times 10^{-5}$ during training. We use the Adam optimizer and train for 30 epochs with a batch size of 8. For VRCM, it is trained for lossy compression using the loss function equation \eqref{eq:loss_all} and \eqref{eq:loss_all_stage3}. We initialize the learning rate to $8 \times 10^{-4}$ and gradually decrease it to $1 \times 10^{-5}$ during training. We use the Adam optimizer and train for 20 epochs with a batch size of 8. We set the $\lambda$ to $0.3$ and $3.0$.

\subsection{Lossless Compression Results}
% \subsection{UELC}
% \input{table/lossless1}
Compared to G-PCC, we achieve a compression ratio (CR)-gain of 19.6\%. Compared to SparsePCGC, we achieve a CR-gain of 5.6\% when training dataset is ShapeNet, as shown in Supplementary Material. 
% Additionally, our encoding and decoding speeds are faster than those of G-PCC and SparsePCGC. 
% Please note that the time tests for encoding and decoding are conducted on RTX 4080 GPU and Intel i5-13600KF CPU. We conduct our own reimplementation of SparsePCGC, and closely match the results reported in the original paper.

To compare with Voxeldnn, Octattention and ECM-OPCC methods, we train UELC using 8iVFB and MVUB and the test results are shown in Figure \ref{fig:xingneng} and Table \ref{table:lossless_8i}.
UELC demonstrates improved performance on the 8iVFB testset compared to SparsePCGC, with a 8.1\% improvement. It also outperforms Voxeldnn \cite{voxeldnn} with a 17.4\% improvement, MsVoxelDNN \cite{msvoxel} with a 35.2\% improvement, Octattention\cite{fu2022octattention} with a 21.2\% improvement, GPCC v23 with a significant 29.2\% improvement and ECM-OPCC \cite{ecm} with a 11.3\% improvement. More detailed compression experiments are in the supplementary material. Compared to the previous SOTA methods, our model achieves a performance gain of 8.1\%. Notably, our model exhibits a small model size \cite{gao2022openpointcloud} and fast inference speed, as shown in Figure \ref{fig:xingneng}.
To evaluate the generalization of UELC, we also use UELC for lidar point cloud compression. 
More experimental results are shown in the Supplementary Material.

\subsection{Lossy Compression Results}
% \subsection{VRCM}
% \input{table/lossy_gpcc}
% Please add the following required packages to your document preamble:
% \usepackage{multirow}
\begin{table*}[]
% \caption{{BD-Rate gains measured using both D1 and D2 for the UniPCGC against the PCGCv2, ADL-PCC, GeoCNNv2 and SparsePCGC for lossy compression.}}
% \label{table:lossy}
\centering
\begin{tabular}{ccccccccc}
\hline
\multicolumn{1}{|c|}{\multirow{2}{*}{\textbf{\begin{tabular}[c]{@{}c@{}}Dense\\ Point Clouds\end{tabular}}}} & \multicolumn{2}{c|}{\textbf{PCGCv2}}                          & \multicolumn{2}{c|}{\textbf{ADL-PCC}}                         & \multicolumn{2}{c|}{\textbf{GeoCNNv2}}                        & \multicolumn{2}{c|}{\textbf{SparsePCGC}}                      \\ \cline{2-9} 
\multicolumn{1}{|c|}{}                                                                                       & \textbf{D1}          & \multicolumn{1}{c|}{\textbf{D2}}       & \textbf{D1}          & \multicolumn{1}{c|}{\textbf{D2}}       & \textbf{D1}          & \multicolumn{1}{c|}{\textbf{D2}}       & \textbf{D1}          & \multicolumn{1}{c|}{\textbf{D2}}       \\ \hline
\multicolumn{1}{|c|}{longdress\_vox10\_1300}                                                                 & -49.76\%             & \multicolumn{1}{c|}{-41.73\%}          & -77.90\%             & \multicolumn{1}{c|}{-76.34\%}          & -75.17\%             & \multicolumn{1}{c|}{-70.19\%}          & -14.15\%             & \multicolumn{1}{c|}{-14.47\%}          \\
\multicolumn{1}{|c|}{loot\_vox10\_1200}                                                                      & -50.31\%             & \multicolumn{1}{c|}{-42.85\%}          & -78.48\%             & \multicolumn{1}{c|}{-76.86\%}          & -76.43\%             & \multicolumn{1}{c|}{-71.27\%}          & -19.95\%              & \multicolumn{1}{c|}{-14.85\%}           \\
\multicolumn{1}{|c|}{red\&black\_vox10\_1550}                                                                & -49.16\%             & \multicolumn{1}{c|}{-36.54\%}          & -71.20\%             & \multicolumn{1}{c|}{-70.03\%}          & -73.07\%             & \multicolumn{1}{c|}{-66.03\%}          & -11.09\%             & \multicolumn{1}{c|}{-9.43\%}          \\
\multicolumn{1}{|c|}{soldier\_vox10\_0690}                                                                   & -48.57\%             & \multicolumn{1}{c|}{-41.20\%}          & -74.28\%             & \multicolumn{1}{c|}{-72.46\%}          & -73.84\%             & \multicolumn{1}{c|}{-68.45\%}          & -15.91\%             & \multicolumn{1}{c|}{-12.63\%}          \\
\multicolumn{1}{|c|}{queen\_vox10\_0200}                                                                            & -41.30\%             & \multicolumn{1}{c|}{-37.77\%}          & -79.07\%             & \multicolumn{1}{c|}{-78.64\%}          & -73.27\%             & \multicolumn{1}{c|}{-67.80\%}          & -12.15\%              & \multicolumn{1}{c|}{-12.79\%}          \\
\multicolumn{1}{|c|}{player\_vox11\_0200}                                                                     & -53.86\%             & \multicolumn{1}{c|}{-48.58\%}          & -83.03\%             & \multicolumn{1}{c|}{-81.79\%}          & -80.77\%             & \multicolumn{1}{c|}{-76.86\%}          & -12.44\%             & \multicolumn{1}{c|}{-11.41\%}          \\
\multicolumn{1}{|c|}{dancer\_vox11\_0001}                                                                    & -53.00\%             & \multicolumn{1}{c|}{-45.32\%}          & -81.56\%             & \multicolumn{1}{c|}{-78.39\%}          & -79.01\%             & \multicolumn{1}{c|}{-72.40\%}          & -12.43\%             & \multicolumn{1}{c|}{-9.80\%}          \\ \hline
\multicolumn{1}{|c|}{\textbf{Average}}                                                                       & \textbf{-49.75\%}    & \multicolumn{1}{c|}{\textbf{-42.00\%}} & \textbf{-77.93\%}    & \multicolumn{1}{c|}{\textbf{-76.36\%}} & \textbf{-75.94\%}    & \multicolumn{1}{c|}{\textbf{-70.43\%}} & \textbf{-14.02\%}    & \multicolumn{1}{c|}{\textbf{-12.20\%}} \\ \hline
\multicolumn{1}{l}{}                                                                                         & \multicolumn{1}{l}{} & \multicolumn{1}{l}{}                   & \multicolumn{1}{l}{} & \multicolumn{1}{l}{}                   & \multicolumn{1}{l}{} & \multicolumn{1}{l}{}                   & \multicolumn{1}{l}{} & \multicolumn{1}{l}{}                   \\
\multicolumn{1}{l}{}                                                                                         & \multicolumn{1}{l}{} & \multicolumn{1}{l}{}                   & \multicolumn{1}{l}{} & \multicolumn{1}{l}{}                   & \multicolumn{1}{l}{} & \multicolumn{1}{l}{}                   & \multicolumn{1}{l}{} & \multicolumn{1}{l}{}                  
\end{tabular}
\caption{{BD-Rate gains measured using both D1 and D2 for the UniPCGC against the PCGCv2, ADL-PCC, GeoCNNv2 and SparsePCGC for lossy compression.}}
\label{table:lossy}
\end{table*}

In the lossy compression, we set $n$ as 0,1,2 to achieve variable rate within different ranges. Due to space limitations, table in the Supplementary Material presents the comparison between our approach and the latest traditional method G-PCC. Using D1 and D2 as distortion metrics, our method shows an average Bjontegaard Delta (BD)-rate improvement of 93.26\% and 88.47\% over the latest G-PCC v23. We also demonstrate superior performance compared to learning-based methods. VRCM outperforms SparsePCGC \cite{sparsepcgc} by 14.02\% and 12.20\% in BD-rate, PCGCv2 \cite{pcgcv2} by 49.75\% and 42.00\% in BD-rate, GeoCNNv2 \cite{geocnnv2} by 75.94\% and 70.43\% in BD-rate, and ADL-PCC \cite{adl_pcc} by 77.93\% and 76.36\% in BD-rate, as shown in Tabel \ref{table:lossy} and Figure \ref{fig:rdcurves}. 
We not only achieve state-of-the-art RD performance but also support variable rate and complexity.

By setting different values for $\lambda$ and $n$, fine-grained rate modulation can be achieved. When $\lambda \in (-20,3)$ and $n = 0,1,2$, continuous rate variation can be achieved. Due to GPU memory limitations (less than 16GB), we use only two values of $\lambda$ during model training, which already encompass a wide range of rates. If more values of $\lambda$ are used during model training, the RD performance is expected to improve further.
Moreover, our model achieves faster inference speed at low rate. Compared to static models, our dynamic network model achieves a encoding acceleration of $59.6\%$ at the lowest rate and a decoding acceleration of $33.4\%$ in loot 1200. We provide RD curves for variable rate and complexity in the Supplementary Material.

\begin{table}[]
% \caption{{The ablation experiments of UELC. %\checkmark denotes the usage of that method.
% }}
% \label{table:ablation}
\centering
\begin{tabular}{|c|ccc|c|c|}
\hline
\textbf{Method}                & \textbf{SG} & \textbf{NSG} & \textbf{UEG} & \textbf{Gain} & \textbf{Parms}  \\ \hline
Baseline           & \checkmark  &              &             & 0.0\%   & 5.0M         \\
Baseline+NSG                   &             & \checkmark   &             & -2.0\%     & 1.3M     \\
Baseline+UE                    &  \checkmark           &              & \checkmark  & -1.8\%  & 2.3M        \\
\textbf{UELC} &             & \checkmark   & \checkmark  & \textbf{-5.6\%} & 2.3M \\ \hline
\end{tabular}
\caption{{The ablation experiments of UELC. %\checkmark denotes the usage of that method.
}}
\label{table:ablation}
\end{table}
\begin{table}[]
% \caption{The ablation experiments of VRCM.} 
% \label{table:ablation_lossy}
\centering
\begin{tabular}{|c|ccc|c|c|}
\hline
\textbf{Method}                & \textbf{UELC} & \textbf{VR} & \textbf{VC} & \textbf{D1-Gain }  & \textbf{D2-Gain}  \\ \hline
Base           &   &              &             & 0.0\%   & 0.0\%        \\
UELC                 &          \checkmark     &   &             & -2.03\%     & -2.09\%     \\
VRM                &  \checkmark           &         \checkmark     &   & -8.20\%  & -7.07\%         \\
\textbf{VRCM} &     \checkmark        & \checkmark   & \checkmark  & \textbf{-14.02\%} & \textbf{-12.20\%}  \\ \hline
\end{tabular}
\caption{The ablation experiments of VRCM.} 
\label{table:ablation_lossy}
\end{table}
\subsection{Ablation Studies}

Table \ref{table:ablation} presents the ablation results of proposed UELC. In UELC, we propose two techniques: Uneven Eight grouping (UEG) and Non-Sequential grouping (NSG). SparsePCGC is used as the baseline, which employs sequential grouping (SG) and uniform 8 grouping scheme. In UEG, a larger distance is considered within each group, so we set $k$ to $5$ for feature extraction. To balance the cost introduced by $k=5$, we set $c$ to $16$ as a trade-off between performance and complexity. UELC combines NSG and UEG, resulting in a parameter count of 2.3M and achieving a $5.6\%$ CR-gain compared to SparsePCGC, which demonstrates the effectiveness of the UEG and NSG schemes. 
Table \ref{table:ablation_lossy} is the ablation experiment of VRCM. D1/D2-Gain represents the BD(D1/D2)-rate gain over the baseline. VR stands for the proposed rate modulation module, and VC stands for the proposed dynamic sparse convolution. 
%When the two technologies are used together, it can bring a 14.02\% BD rate gain, which proves the effectiveness of the proposed module.
Combining the baseline with UELC results in BD-rate gains of 2.03\% and 2.09\%. Building upon this, the introduction of the variable rate control module not only brings about BD-rate gains of 8.20\% and 7.07\%, but also enables the model to have the capability of variable rate encoding. On the basis of UELC and VR, further gains of 14.02\% and 12.20\% in BD-rate are achieved by introducing dynamic sparse convolution, which proves the effectiveness of the proposed module.

\section{Conclusion}
The paper presents an efficient unified framework named UniPCGC, which supports lossless compression, lossy compression, variable rate and variable complexity. 
First, we introduce the UELC in the lossless mode, which allocates more computational complexity to groups with higher coding difficulty, and merges groups with lower coding difficulty. Second, VRCM is achieved in the lossy mode through joint adoption of a rate modulation module and dynamic sparse convolution. 
Compared with the previous state-of-the-art works, the proposed UniPCGC achieves a CR-gain of 8.1\% on lossless compression, a BD-Rate gain of 14.02\% on lossy compression, and supports variable rate and complexity. 
These advancement promote the flexibility and practicality of point cloud geometry compression.

\section{Acknowledgments}
This work was supported by The Major Key Project of PCL (PCL2024A02), Natural Science Foundation of China (62271013, 62031013), Guangdong Provincial Key Laboratory of Ultra High Definition Immersive Media Technology (2024B1212010006), Guangdong Province Pearl River Talent Program (2021QN020708), Guangdong Basic and Applied Basic Research Foundation (2024A1515010155), Shenzhen Science and Technology Program (JCYJ20240813160202004, JCYJ20230807120808017).

% The preparation of the \LaTeX{} and Bib\TeX{} files that implement these instructions was supported by Schlumberger Palo Alto Research, AT\&T Bell Laboratories, Morgan Kaufmann Publishers, The Live Oak Press, LLC, and AAAI Press. Bibliography style changes were added by Sunil Issar. \verb+\+pubnote was added by J. Scott Penberthy. George Ferguson added support for printing the AAAI copyright slug. Additional changes to aaai25.sty and aaai25.bst have been made by Francisco Cruz and Marc Pujol-Gonzalez.

% \bigskip
% \noindent Thank you for reading these instructions carefully. We look forward to receiving your electronic files!

\nocite{*}
\bibliography{aaai25}

\begin{thebibliography}{43}
\providecommand{\natexlab}[1]{#1}

\bibitem[{Ali et~al.(2024)Ali, Kim, Qamar, Lim, Kim, Zhang, Bae, and Kim}]{ali2024towards}
Ali, M.~S.; Kim, Y.; Qamar, M.; Lim, S.-C.; Kim, D.; Zhang, C.; Bae, S.-H.; and Kim, H.~Y. 2024.
\newblock Towards efficient image compression without autoregressive models.
\newblock \emph{Advances in Neural Information Processing Systems}, 36.

\bibitem[{Ball{\'e} et~al.(2018)Ball{\'e}, Minnen, Singh, Hwang, and Johnston}]{balle2018variational}
Ball{\'e}, J.; Minnen, D.; Singh, S.; Hwang, S.~J.; and Johnston, N. 2018.
\newblock Variational image compression with a scale hyperprior.
\newblock In \emph{International Conference on Learning Representations}.

\bibitem[{Biswas et~al.(2020)Biswas, Liu, Wong, Wang, and Urtasun}]{biswas2020muscle}
Biswas, S.; Liu, J.; Wong, K.; Wang, S.; and Urtasun, R. 2020.
\newblock Muscle: Multi sweep compression of lidar using deep entropy models.
\newblock \emph{Advances in Neural Information Processing Systems}, 33: 22170--22181.

\bibitem[{Bolukbasi et~al.(2017)Bolukbasi, Wang, Dekel, and Saligrama}]{37}
Bolukbasi, T.; Wang, J.; Dekel, O.; and Saligrama, V. 2017.
\newblock Adaptive neural networks for efficient inference.
\newblock In \emph{International Conference on Machine Learning}, 527--536. PMLR.

\bibitem[{Cao et~al.(2021)Cao, Preda, Zakharchenko, Jang, and Zaharia}]{MPEG_PCC_PIEEE}
Cao, C.; Preda, M.; Zakharchenko, V.; Jang, E.~S.; and Zaharia, T. 2021.
\newblock Compression of Sparse and Dense Dynamic Point Clouds—Methods and Standards.
\newblock \emph{Proceedings of the IEEE}, 109(9): 1537--1558.

\bibitem[{Chang et~al.(2015)Chang, Funkhouser, Guibas, Hanrahan, Huang, Li, Savarese, Savva, Song, Su et~al.}]{chang2015shapenet}
Chang, A.~X.; Funkhouser, T.; Guibas, L.; Hanrahan, P.; Huang, Q.; Li, Z.; Savarese, S.; Savva, M.; Song, S.; Su, H.; et~al. 2015.
\newblock Shapenet: An information-rich 3d model repository.
\newblock \emph{arXiv preprint arXiv:1512.03012}.

\bibitem[{Choi, El-Khamy, and Lee(2019)}]{24}
Choi, Y.; El-Khamy, M.; and Lee, J. 2019.
\newblock Variable rate deep image compression with a conditional autoencoder.
\newblock In \emph{Proceedings of the IEEE/CVF International Conference on Computer Vision}, 3146--3154.

\bibitem[{Choy, Gwak, and Savarese(2019)}]{sparseconv}
Choy, C.; Gwak, J.; and Savarese, S. 2019.
\newblock 4d spatio-temporal convnets: Minkowski convolutional neural networks.
\newblock In \emph{Proceedings of the IEEE/CVF conference on computer vision and pattern recognition}, 3075--3084.

\bibitem[{Cui et~al.(2021)Cui, Wang, Gao, Guo, Feng, and Bai}]{26}
Cui, Z.; Wang, J.; Gao, S.; Guo, T.; Feng, Y.; and Bai, B. 2021.
\newblock Asymmetric gained deep image compression with continuous rate adaptation.
\newblock In \emph{Proceedings of the IEEE/CVF Conference on Computer Vision and Pattern Recognition}, 10532--10541.

\bibitem[{Fu et~al.(2022)Fu, Li, Song, Gao, and Liu}]{fu2022octattention}
Fu, C.; Li, G.; Song, R.; Gao, W.; and Liu, S. 2022.
\newblock Octattention: Octree-based large-scale contexts model for point cloud compression.
\newblock In \emph{Proceedings of the AAAI conference on artificial intelligence}, volume~36, 625--633.

\bibitem[{Gao et~al.(2022)Gao, Ye, Li, Zheng, Wu, and Xie}]{gao2022openpointcloud}
Gao, W.; Ye, H.; Li, G.; Zheng, H.; Wu, Y.; and Xie, L. 2022.
\newblock OpenPointCloud: An open-source algorithm library of deep learning based point cloud compression.
\newblock In \emph{Proceedings of the 30th ACM international conference on multimedia}, 7347--7350.

\bibitem[{Graves(2016)}]{38}
Graves, A. 2016.
\newblock Adaptive computation time for recurrent neural networks.
\newblock \emph{arXiv preprint arXiv:1603.08983}.

\bibitem[{He et~al.(2021)He, Zheng, Sun, Wang, and Qin}]{checkerboard}
He, D.; Zheng, Y.; Sun, B.; Wang, Y.; and Qin, H. 2021.
\newblock Checkerboard context model for efficient learned image compression.
\newblock In \emph{Proceedings of the IEEE/CVF Conference on Computer Vision and Pattern Recognition}, 14771--14780.

\bibitem[{Huang et~al.(2017)Huang, Chen, Li, Wu, Van Der~Maaten, and Weinberger}]{35}
Huang, G.; Chen, D.; Li, T.; Wu, F.; Van Der~Maaten, L.; and Weinberger, K.~Q. 2017.
\newblock Multi-scale dense networks for resource efficient image classification.
\newblock \emph{arXiv preprint arXiv:1703.09844}.

\bibitem[{Huang et~al.(2018)Huang, Liu, Van~der Maaten, and Weinberger}]{32}
Huang, G.; Liu, S.; Van~der Maaten, L.; and Weinberger, K.~Q. 2018.
\newblock Condensenet: An efficient densenet using learned group convolutions.
\newblock In \emph{Proceedings of the IEEE conference on computer vision and pattern recognition}, 2752--2761.

\bibitem[{Huang et~al.(2020)Huang, Wang, Wong, Liu, and Urtasun}]{huang2020octsqueeze}
Huang, L.; Wang, S.; Wong, K.; Liu, J.; and Urtasun, R. 2020.
\newblock Octsqueeze: Octree-structured entropy model for lidar compression.
\newblock In \emph{Proceedings of the IEEE/CVF conference on computer vision and pattern recognition}, 1313--1323.

\bibitem[{Jia et~al.(2022)Jia, Ge, Wang, Ma, and Gao}]{28}
Jia, C.; Ge, Z.; Wang, S.; Ma, S.; and Gao, W. 2022.
\newblock Rate distortion characteristic modeling for neural image compression.
\newblock In \emph{2022 Data Compression Conference (DCC)}, 202--211. IEEE.

\bibitem[{Jin et~al.(2024)Jin, Zhu, Xu, Lin, and Wang}]{ecm}
Jin, Y.; Zhu, Z.; Xu, T.; Lin, Y.; and Wang, Y. 2024.
\newblock ECM-OPCC: Efficient Context Model for Octree-Based Point Cloud Compression.
\newblock In \emph{ICASSP 2024 - 2024 IEEE International Conference on Acoustics, Speech and Signal Processing (ICASSP)}, 7985--7989.

\bibitem[{Lin et~al.(2021)Lin, Liu, Liang, Li, and Wu}]{22}
Lin, J.; Liu, D.; Liang, J.; Li, H.; and Wu, F. 2021.
\newblock A deeply modulated scheme for variable-rate video compression.
\newblock In \emph{2021 IEEE International Conference on Image Processing (ICIP)}, 3722--3726. IEEE.

\bibitem[{Liu et~al.(2017)Liu, Li, Shen, Huang, Yan, and Zhang}]{34}
Liu, Z.; Li, J.; Shen, Z.; Huang, G.; Yan, S.; and Zhang, C. 2017.
\newblock Learning efficient convolutional networks through network slimming.
\newblock In \emph{Proceedings of the IEEE international conference on computer vision}, 2736--2744.

\bibitem[{Nguyen and Kaup(2022)}]{sparsevoxeldnn}
Nguyen, D.~T.; and Kaup, A. 2022.
\newblock Learning-based lossless point cloud geometry coding using sparse tensors.
\newblock In \emph{2022 IEEE International Conference on Image Processing (ICIP)}, 2341--2345. IEEE.

\bibitem[{Nguyen and Kaup(2023)}]{cnet}
Nguyen, D.~T.; and Kaup, A. 2023.
\newblock Lossless point cloud geometry and attribute compression using a learned conditional probability model.
\newblock \emph{IEEE Transactions on Circuits and Systems for Video Technology}.

\bibitem[{Nguyen et~al.(2021{\natexlab{a}})Nguyen, Quach, Valenzise, and Duhamel}]{voxeldnn}
Nguyen, D.~T.; Quach, M.; Valenzise, G.; and Duhamel, P. 2021{\natexlab{a}}.
\newblock Learning-based lossless compression of 3d point cloud geometry.
\newblock In \emph{ICASSP 2021-2021 IEEE International Conference on Acoustics, Speech and Signal Processing (ICASSP)}, 4220--4224. IEEE.

\bibitem[{Nguyen et~al.(2021{\natexlab{b}})Nguyen, Quach, Valenzise, and Duhamel}]{msvoxel}
Nguyen, D.~T.; Quach, M.; Valenzise, G.; and Duhamel, P. 2021{\natexlab{b}}.
\newblock Multiscale deep context modeling for lossless point cloud geometry compression.
\newblock \emph{arXiv preprint arXiv:2104.09859}.

\bibitem[{Pang, Lodhi, and Tian(2022)}]{pang2022grasp}
Pang, J.; Lodhi, M.~A.; and Tian, D. 2022.
\newblock GRASP-Net: Geometric residual analysis and synthesis for point cloud compression.
\newblock In \emph{Proceedings of the 1st International Workshop on Advances in Point Cloud Compression, Processing and Analysis}, 11--19.

\bibitem[{Pereira(2021)}]{adl_pcc}
Pereira, A. G. N. R.~F. 2021.
\newblock Adaptive Deep Learning-Based Point Cloud Geometry Coding.
\newblock \emph{IEEE Journal on Selected Topics in Signal Processing}, 15: 415--430.

\bibitem[{Quach, Valenzise, and Dufaux(2020)}]{geocnnv2}
Quach, M.; Valenzise, G.; and Dufaux, F. 2020.
\newblock Improved Deep Point Cloud Geometry Compression.
\newblock \emph{2020 IEEE MMSP Workshop}.

\bibitem[{Song, Choi, and Han(2021)}]{21}
Song, M.; Choi, J.; and Han, B. 2021.
\newblock Variable-rate deep image compression through spatially-adaptive feature transform.
\newblock In \emph{Proceedings of the IEEE/CVF International Conference on Computer Vision}, 2380--2389.

\bibitem[{Song et~al.(2023)Song, Fu, Liu, and Li}]{song2023efficient}
Song, R.; Fu, C.; Liu, S.; and Li, G. 2023.
\newblock Efficient Hierarchical Entropy Model for Learned Point Cloud Compression.
\newblock In \emph{Proceedings of the IEEE/CVF Conference on Computer Vision and Pattern Recognition}, 14368--14377.

\bibitem[{Sun et~al.(2021)Sun, Tan, Sun, Zhang, Qian, Li, and Li}]{20}
Sun, Z.; Tan, Z.; Sun, X.; Zhang, F.; Qian, Y.; Li, D.; and Li, H. 2021.
\newblock Interpolation variable rate image compression.
\newblock In \emph{Proceedings of the 29th ACM international conference on multimedia}, 5574--5582.

\bibitem[{Tao and Gao(2021)}]{taoecp}
Tao, L.; and Gao, W. 2021.
\newblock Efficient Channel Pruning Based on Architecture Alignment and Probability Model Bypassing.
\newblock In \emph{2021 IEEE International Conference on Systems, Man, and Cybernetics (SMC)}, 3232--3237.

\bibitem[{Tao et~al.(2023)Tao, Gao, Li, and Zhang}]{tao2023adanic}
Tao, L.; Gao, W.; Li, G.; and Zhang, C. 2023.
\newblock Adanic: Towards practical neural image compression via dynamic transform routing.
\newblock In \emph{Proceedings of the IEEE/CVF International Conference on Computer Vision}, 16879--16888.

\bibitem[{Teerapittayanon, McDanel, and Kung(2016)}]{36}
Teerapittayanon, S.; McDanel, B.; and Kung, H.-T. 2016.
\newblock Branchynet: Fast inference via early exiting from deep neural networks.
\newblock In \emph{2016 23rd international conference on pattern recognition (ICPR)}, 2464--2469. IEEE.

\bibitem[{Theis et~al.(2016)Theis, Shi, Cunningham, and Husz{\'a}r}]{27}
Theis, L.; Shi, W.; Cunningham, A.; and Husz{\'a}r, F. 2016.
\newblock Lossy image compression with compressive autoencoders.
\newblock In \emph{International Conference on Learning Representations}.

\bibitem[{Veit and Belongie(2018)}]{40}
Veit, A.; and Belongie, S. 2018.
\newblock Convolutional networks with adaptive inference graphs.
\newblock In \emph{Proceedings of the European conference on computer vision (ECCV)}, 3--18.

\bibitem[{Wang et~al.(2022)Wang, Ding, Li, Feng, Cao, and Ma}]{sparsepcgc}
Wang, J.; Ding, D.; Li, Z.; Feng, X.; Cao, C.; and Ma, Z. 2022.
\newblock Sparse tensor-based multiscale representation for point cloud geometry compression.
\newblock \emph{IEEE Transactions on Pattern Analysis and Machine Intelligence}.

\bibitem[{Wang et~al.(2021)Wang, Ding, Li, and Ma}]{pcgcv2}
Wang, J.; Ding, D.; Li, Z.; and Ma, Z. 2021.
\newblock Multiscale point cloud geometry compression.
\newblock In \emph{2021 Data Compression Conference (DCC)}, 73--82. IEEE.

\bibitem[{Wang et~al.(2019)Wang, Zhu, Ma, Chen, Liu, and Shen}]{pcgcv1}
Wang, J.; Zhu, H.; Ma, Z.; Chen, T.; Liu, H.; and Shen, Q. 2019.
\newblock Learned point cloud geometry compression.
\newblock \emph{arXiv preprint arXiv:1909.12037}.

\bibitem[{Wang et~al.(2018)Wang, Yu, Dou, Darrell, and Gonzalez}]{39}
Wang, X.; Yu, F.; Dou, Z.-Y.; Darrell, T.; and Gonzalez, J.~E. 2018.
\newblock Skipnet: Learning dynamic routing in convolutional networks.
\newblock In \emph{Proceedings of the European conference on computer vision (ECCV)}, 409--424.

\bibitem[{Xie et~al.(2024)Xie, Gao, Zheng, and Li}]{xie2024roi}
Xie, L.; Gao, W.; Zheng, H.; and Li, G. 2024.
\newblock ROI-Guided Point Cloud Geometry Compression Towards Human and Machine Vision.
\newblock In \emph{Proceedings of the 32nd ACM International Conference on Multimedia}.

\bibitem[{Yang et~al.(2020)Yang, Herranz, Van De~Weijer, Guiti{\'a}n, L{\'o}pez, and Mozerov}]{23}
Yang, F.; Herranz, L.; Van De~Weijer, J.; Guiti{\'a}n, J. A.~I.; L{\'o}pez, A.~M.; and Mozerov, M.~G. 2020.
\newblock Variable rate deep image compression with modulated autoencoder.
\newblock \emph{IEEE Signal Processing Letters}, 27: 331--335.

\bibitem[{Yin et~al.(2022)Yin, Li, Bao, Liang, Meng, and Liu}]{25}
Yin, S.; Li, C.; Bao, Y.; Liang, Y.; Meng, F.; and Liu, W. 2022.
\newblock Universal efficient variable-rate neural image compression.
\newblock In \emph{ICASSP 2022-2022 IEEE International Conference on Acoustics, Speech and Signal Processing (ICASSP)}, 2025--2029. IEEE.

\bibitem[{You et~al.(2024)You, Liu, Yu, Gao, and Ding}]{ijcai2024p595}
You, K.; Liu, K.; Yu, L.; Gao, P.; and Ding, D. 2024.
\newblock Pointsoup: High-Performance and Extremely Low-Decoding-Latency Learned Geometry Codec for Large-Scale Point Cloud Scenes.
\newblock In Larson, K., ed., \emph{Proceedings of the Thirty-Third International Joint Conference on Artificial Intelligence, {IJCAI-24}}, 5380--5388. International Joint Conferences on Artificial Intelligence Organization.
\newblock Main Track.

\end{thebibliography}

\end{document}